\pdfoutput=1
\def\year{2021}\relax
\documentclass[letterpaper]{article} 
\usepackage{aaai21}  
\usepackage{times}  
\usepackage{helvet} 
\usepackage{courier}  
\usepackage[hyphens]{url}  
\usepackage{graphicx} 
\usepackage{natbib}  
\usepackage{caption} 
\usepackage[flushleft]{threeparttable}
\usepackage[bottom]{footmisc}
\usepackage[noend]{algcompatible}
\usepackage[noend]{algpseudocode}
\usepackage{epsfig}
\usepackage{fancyhdr}
\usepackage[normalem]{ulem}
\usepackage{amsmath}
\usepackage{multirow}
\usepackage{caption}
\usepackage{float}
\usepackage{amssymb}
\usepackage{mathtools}
\usepackage{xspace}
\usepackage{color}
\usepackage{comment}
\usepackage{graphicx}
\usepackage{pifont}
\usepackage{listings}
\usepackage{subdepth}
\usepackage{subcaption}
\usepackage{algorithm}
\usepackage{dsfont}
\usepackage{setspace}
\usepackage{lipsum}
\usepackage[table,prologue]{xcolor}
\usepackage{todonotes}
\usepackage[switch]{lineno}  %

\newcommand{\bheading}[1]{{\vspace{2pt}\noindent{\textbf{#1}}\hspace{2pt}}}

\title{ActionBert: Leveraging User Actions for Semantic Understanding of User Interfaces}
\author {
    Zecheng He \textsuperscript{\rm 1},
    Srinivas Sunkara \textsuperscript{\rm 2},
    Xiaoxue Zang \textsuperscript{\rm 2},
    Ying Xu \textsuperscript{\rm 2},
    Lijuan Liu \textsuperscript{\rm 2},
    Nevan Wichers \textsuperscript{\rm 2},
    Gabriel Schubiner  \textsuperscript{\rm 2},
    Ruby Lee \textsuperscript{\rm 1},
    Jindong Chen \textsuperscript{\rm 2},
    Blaise Ag\"{u}era y Arcas \textsuperscript{\rm 2} \\
}

\affiliations {
    \textsuperscript{\rm 1} Princeton University
    \textsuperscript{\rm 2} Google Research \\
    \{zechengh, rblee\}@princeton.edu, \{srinivasksun, xiaoxuez, yingyingxuxu, lijuanliu, wichersn, gsch, jdchen, blaisea\}@google.com
}

\begin{document}
\maketitle

\begin{abstract}

As mobile devices are becoming ubiquitous, regularly interacting with a variety of user interfaces (UIs) is a common aspect of daily life for many people. To improve the accessibility of these devices and to enable their usage in a variety of settings, building models that can assist users and accomplish tasks through the UI is vitally important. However, there are several challenges to achieve this. First, UI components of similar appearance can have different functionalities, making understanding their function more important than just analyzing their appearance. Second, domain-specific features like Document Object Model (DOM) in web pages and View Hierarchy (VH) in mobile applications provide important signals about the semantics of UI elements, but these features are not in a natural language format. Third, owing to a large diversity in UIs and absence of standard DOM or VH representations, building a UI understanding model with high coverage requires large amounts of training data.

Inspired by the success of pre-training based approaches in NLP for tackling a variety of problems in a data-efficient way, we introduce a new pre-trained UI representation model called ActionBert. Our methodology is designed to leverage visual, linguistic and domain-specific features in user interaction traces to pre-train generic feature representations of UIs and their components. Our key intuition is that user actions, e.g., a sequence of clicks on different UI components, reveals important information about their functionality. We evaluate the proposed model on a wide variety of downstream tasks, ranging from icon classification to UI component retrieval based on its natural language description. Experiments show that the proposed ActionBert model outperforms multi-modal baselines across all downstream tasks by up to 15.5\%.

\end{abstract}

\section{Introduction}\label{sec:intro}

 Given the prevalence and importance of smart devices in our daily life, the ability to understand and operate User Interfaces (UIs) has become an important task for Artificial Intelligence. For instance, a model that can find a UI component by its description can be very useful for voice interfaces, and a model that can predict the expected output of clicking a button can help page navigation. To successfully operate a UI, the models need to understand the user task and intents, and how to perform the tasks in the given UI.



However, UI understanding is a challenging and less-studied area. First, there are various tasks related to UI understanding. Usually, these tasks are cross-modal and cross-domain, e.g., clicking a button through voice command and retrieving an icon via a semantically similar one. Previous works in this field usually target one single task at a time. Training a different complex model for each task is not efficient for on-device model deployment. Moreover, models may suffer from overfitting if the task-specific data is limited. Pre-training models on large-scale datasets to extract features has shown great power in multiple domains, e.g., ResNet \cite{he2016deep} in computer vision and BERT \cite{devlin2018bert} in natural language processing. There is no such generic feature representation for user interfaces and it is not clear if a pre-trained feature extraction model can help improve multiple UI related tasks.

Second, the data source and format of UIs are different from  natural image and language corpuses. For example, the View Hierarchy (VH) in mobile apps and Document Object Model (DOM) in web pages are tree structures representing the UI layout. The VH and DOM contain structural and semantic information about the UI, however they are not generally visible to the users and they also contain short phrases with hints about functionality. Effectively making use of this domain-specific knowledge for general UI understanding is an unsolved problem.

\begin{figure}[t]
    \centering
    \includegraphics[width=0.6\columnwidth]{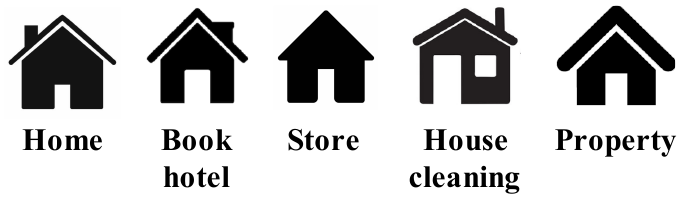}
    \caption{Examples where low-level appearance does not reflect the semantics of UI components without context. Shown below are their semantics on the source screen.}\label{fig:icon-semantic}
\end{figure}

Third, understanding the functionality of UIs is more challenging than learning about their appearance. It is common that UI elements of similar appearance have very different semantic meanings, and vice versa. In the example in Figure \ref{fig:icon-semantic}, all The icons look similar to each other (``houses''), however, they have different functionalities which can only be interpreted with additional context.

In this paper, we propose ActionBert, a pre-trained transformer-style \cite{vaswani2017attention} model that leverages sequential user action information for UI understanding.
Our key intuition is that the semantic meaning of a UI, and the functionality of UI components can be captured by user actions, e.g. a sequence of clicks and their effect on the UI. This model takes advantage of the representation power of transformer models, and integrates domain-specific information like  VH and user actions to build embeddings reflecting the functionality of different UI elements. To the best of our knowledge, this is the first attempt to build a generic feature representation in this field. Our main contributions in this paper are:

\begin{itemize}
\item To the best of our knowledge, we are the first to integrate the powerful transformer models and domain-specific knowledge, e.g., VH and user actions, to improve machines' understanding of UIs.

\item We propose ActionBert, a transformer-style multi-modal model to capture the context and semantic meaning of UI elements by introducing new self-supervised pre-training tasks based on user actions and UI-specific features.


\item We evaluate the proposed model on four types of UI downstream tasks, capturing various real-life use cases. We show that the proposed model outperforms the existing models on all tasks.
\end{itemize}


\section{Background of UI View Hierarchy} \label{sec:bg}

A View hierarchy is a tree-based representation of a user interface. It has various attributes related to the appearance and functionality of UI components. In this paper, we leverage the content description, resource id, component class, component text fields in the leaf nodes of view hierarchy:
\begin{itemize}\setlength{\itemsep}{0pt}
\item The content description is a brief description of the functionality of this UI component provided by the developer.
\item The component text is the visible text on the component.
\item The resource id and component class indicate the type of the component, e.g., button, checkbox or image, and the name of the static files used to render it.
\end{itemize}

\begin{figure}[ht]
    \centering
    \includegraphics[width=0.9\columnwidth]{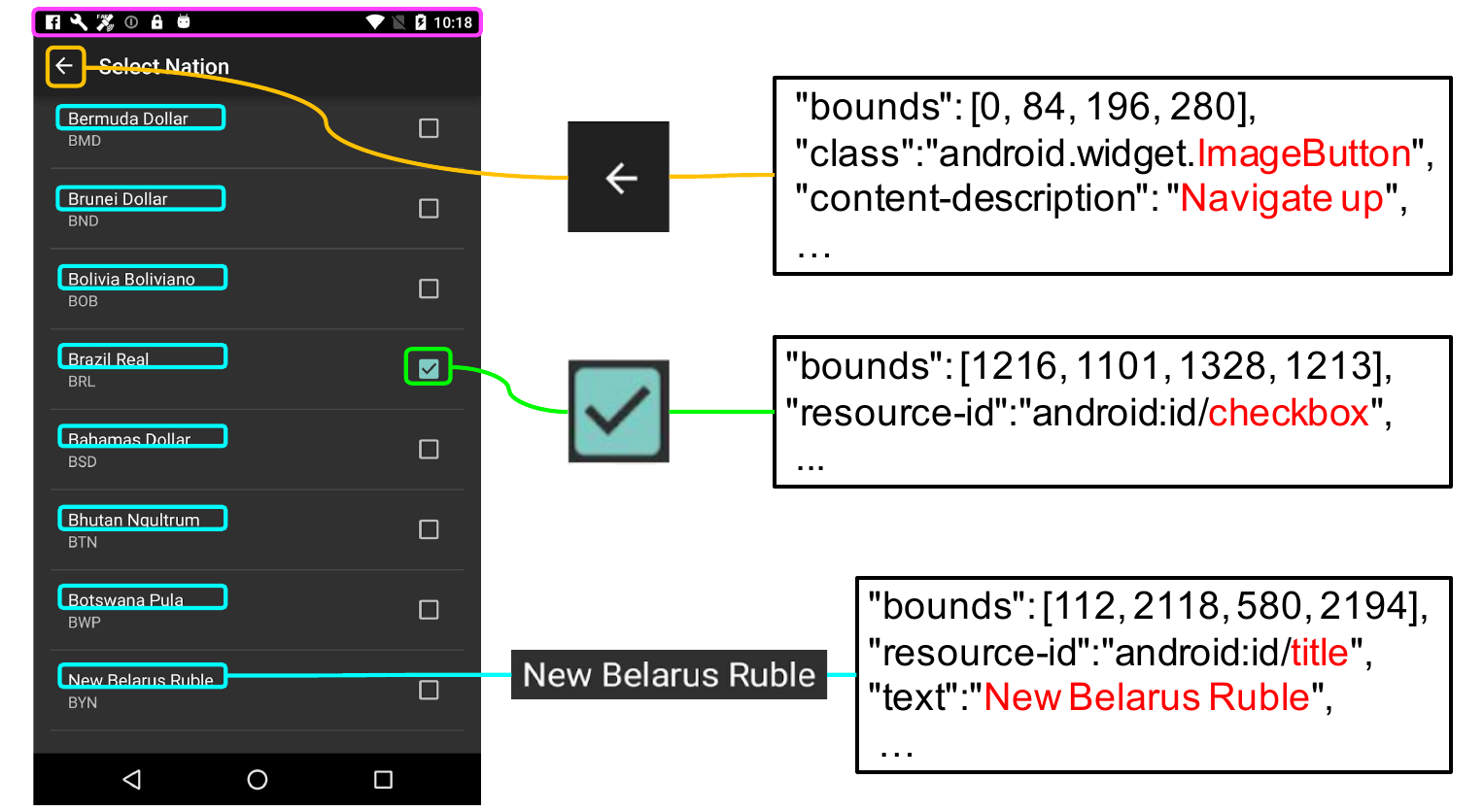}
    \caption{An example of leaf nodes in a view hierarchy. The view hierarchy provides useful semantic information (marked red) for machines to understand the UI.}\label{fig:bg:vh}
\end{figure}

  Some of the important information, like content description, is invisible to the user but can be used by applications like Screen Readers to understand the UI. Figure \ref{fig:bg:vh} shows examples of leaf nodes in a view hierarchy.

\section{ActionBert} \label{sec:actionbert}

\subsection{Revisiting BERT}

ActionBert is inspired by the great success of BERT \cite{devlin2018bert} in natural language processing (NLP). We briefly review the BERT model, and then extend the concepts to learn UI embedding.

BERT is a transformer-based \cite{vaswani2017attention} bidirectional language model. BERT-style models have shown great success in transferring learned features to multiple NLP tasks. On a high level, BERT takes in the embedding of word tokens and processes them through a multi-layer bidirectional transformer \cite{vaswani2017attention}.

\begin{figure}[ht]
    \centering
    \includegraphics[width=0.7\columnwidth]{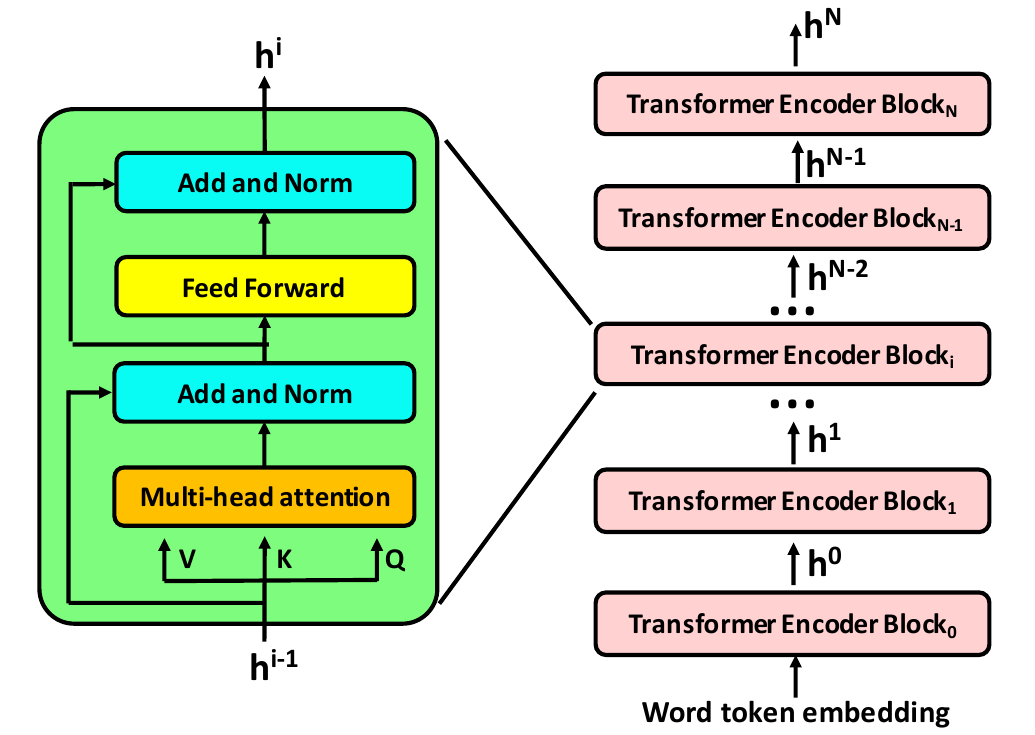}
    \caption{BERT and transformer encoder blocks.}\label{fig:actionbert:transformer-block}
\end{figure}

BERT is pre-trained with two tasks: masked language modeling (MLM) and next sentence prediction (NSP). In MLM task, some input words are randomly masked out and replaced with a special token [MASK]. The task is to predict the masked word based on the clues from the unmasked words in its context. The NSP task is defined as, given two sentences predict whether one is immediately after the other. To separate the two sentences, a special token [SEP] is inserted between them. A classifier is applied to the BERT embedding and outputs the probability of the second sentence immediately following the first one. More details on BERT can be found in \cite{devlin2018bert}.

\subsection{ActionBert: Semantic UI understanding with user actions}

Inspired by the BERT model, we adopt the concepts of NLP and extend them to UI understanding. We treat the UI components, e.g., buttons, icons, checkboxes etc as the basic building blocks of a user interface. Similar to sentences, which are composed of word tokens, we treat these basic UI components (buttons, icons, etc.) as tokens, and the whole UI as a sentence in NLP. A user interaction trace is a sequence of UIs obtained by starting from a particular UI and interacting with different UI components. Different from sentences, UIs in this trace are linked through a link component, usually a clickable component like a button or an icon. When a user takes an action on that link component, the screen jumps to the next UI. Such a sequence of UIs is analogous to paragraphs or documents in language modeling. Table \ref{tab:concepts-map} shows a mapping of the concepts between NLP and UI understanding.

\begin{table}[h]
\centering
\caption{Concepts mapping between NLP and UI understanding.} \label{tab:concepts-map}
\resizebox{\linewidth}{!}{
\begin{tabular}{|c|c|}
\hline
\textbf{Natural Language Processing} & \textbf{UI Understanding}                 \\ \hline
Tokens                               & UI components (buttons, icons, texts etc) \\ \hline
Sentences                            & UIs                                       \\ \hline
Word context                         & UI components in the same UI              \\ \hline
Consecutive sentences                & Consecutive UIs                           \\ \hline
Paragraph/document                   & Sequence of UIs                           \\ \hline
Language model                       & UI embedding model                        \\ \hline
\end{tabular}
}
\end{table}

\begin{figure}[ht]
    \centering
    \includegraphics[width=0.9\columnwidth]{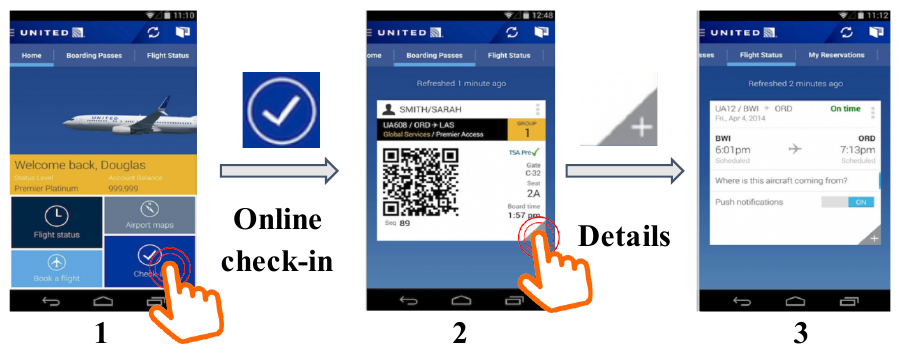}
    \caption{An example of user actions on UIs. The user clicks on the ``tick'' button in screen 1 and jumps to the boarding-pass page, screen 2. The semantics of the button, online check-in, can be inferred from components on the homepage (e.g., images of plane, airline name) and the components on the next UI (e.g., QR code, passenger information).}\label{fig:actionbert:toy-example}
\end{figure}

Following this analogy, our key idea is that the semantic meaning of a UI component can be identified from components in the same UI and the UI that follows the current one. We illustrate this idea in an example in Figure \ref{fig:actionbert:toy-example}. Here, the first UI is the homepage of an airline app. The user clicks the button with a tick and a circle on it. This button links to a new UI with passenger, time, gate information and a QR code on it. From the elements in the current UI and the next UI, the functionality of the button can be interpreted as ``online check-in''. Similarly, when the user clicks on the ``plus'' button it links to a UI with more detailed flight information on it. Hence, the ``plus'' component indicates ``show details''.

\begin{figure*}[ht]
    \centering
    \includegraphics[width=0.85\linewidth]{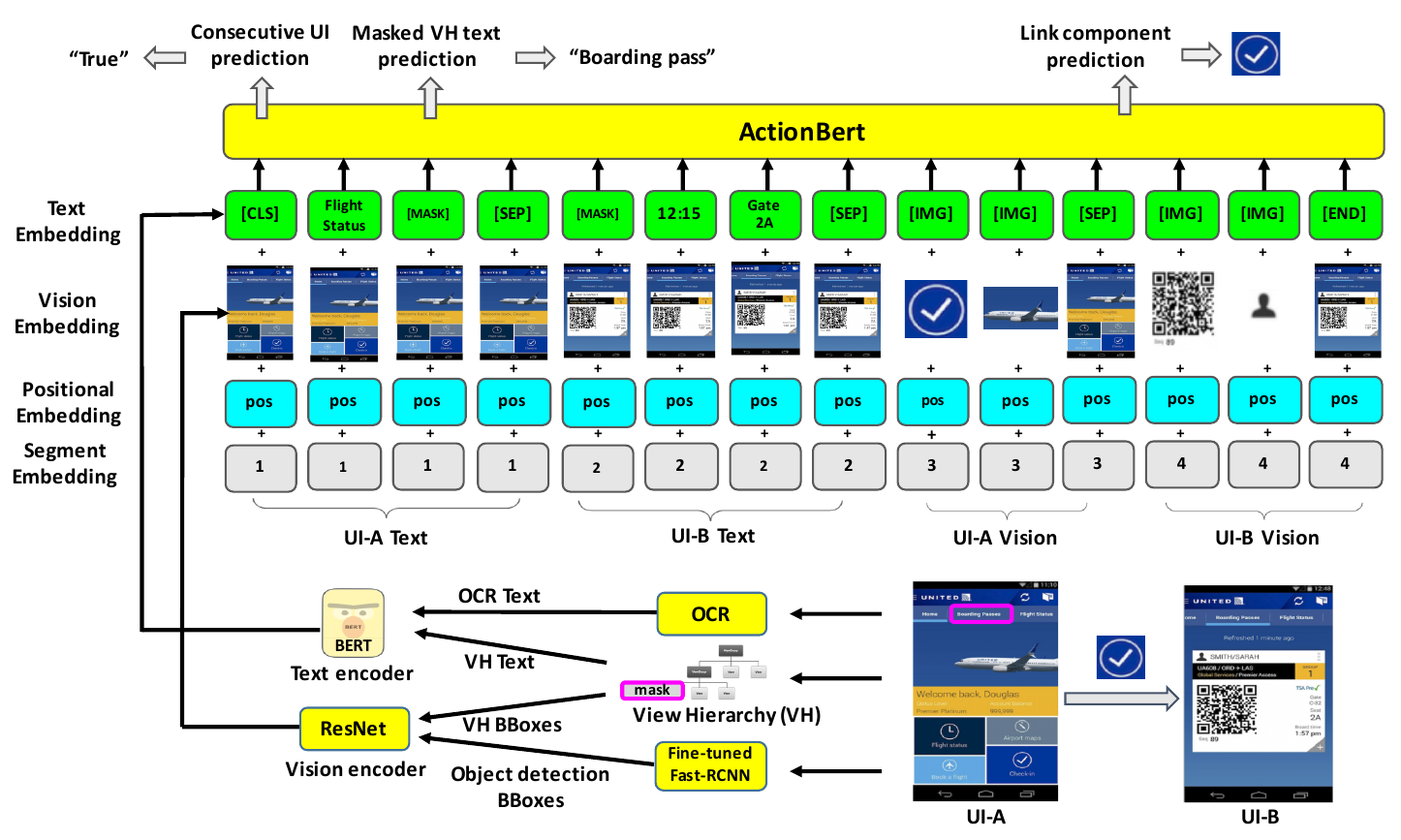}
    \caption{ActionBert model architecture. On a high-level, the ActionBert model takes a pair of UIs, represented by their text, vision, positional and segment embedding as input. Three new UI-specific tasks, i.e. link component prediction, consecutive UI prediction and masked text prediction, are defined to pre-train ActionBert on large-scale UI sequences with user actions.}\label{fig:actionbert:architecture}
\end{figure*}

We propose ActionBert that takes a pair of UIs as input, and outputs the contextual embedding of the UIs and the individual UI components. Figure \ref{fig:actionbert:architecture} shows the model architecture. It extends the original BERT model by adding vision modality and leverages user-action related tasks for pre-training. Inspired by the recent vision-language model VL-BERT \cite{su2019vl}, the ActionBert model uses a uni-stream architecture that allows full-attention across modalities. First, the two input UIs (UI-A and UI-B) are split into four component segments: UI-A text, UI-B text, UI-A vision and UI-B vision. A special token [CLS] is prepended to the component sequence, similar to the original BERT model, whose embedding represents the whole two input UIs. The different segments, representing the text and vision parts of the two UIs are separated with a special token [SEP] and end with another special token [END].

\subsubsection{Text embedding} Different from BERT and other vision-language models, the text tokens of ActionBert are specifically designed for UI tasks. Each text token (green box in Figure \ref{fig:actionbert:architecture}) is a concatenation of content description, resource id, component class name and text in a view hierarchy leaf node (Section \ref{sec:bg}). The vision segment slots of text tokens are filled with a special token [IMG]. Overall, each text token, which is a concatenation of the different fields in View Hierarchy, is treated as a sentence and processed through a sentence-level text encoder, e.g. BERT, to generate the input text embedding.

\subsubsection{Vision embedding} Similar to the text tokens, the vision tokens are also specific to the nature of UIs. If the view hierarchy of a UI is available, each vision token is cropped from the UI using the bounding box of a VH leaf node. If the VH is not available, we fine-tune a Faster-RCNN \cite{ren2015frcnn} to detect UI components in a screenshot and crop components from the detected bounding boxes. Furthermore, a vision encoder, e.g., ResNet-50, is used to generate the input vision embedding from the cropped images. Specifically, from the vision encoder, we take the flattened feature map from the layer just before the fully connected layer as the input vision embedding. Vision tokens of UI-A text, UI-B text and special tokens ([CLS], [SEP] and [END]) are set as the corresponding whole UI screenshots.

\subsubsection{Positional embedding} Positional embedding represents the geometrical position of UI components in the UI. Unlike word tokens in language, components in a UI are not linearly ordered, but are arranged in a 2D space. We define nine features to represent the positional features of a UI component, i.e. $x_{min}$, $y_{min}$, $x_{max}$, $y_{max}$, $x_{center}, y_{center}$, height, width and area. $x_{min}$, $y_{min}$ correspond to the top-left corner and $x_{max}$, $y_{max}$ correspond to the bottom-right corner of the UI component, respectively. To deal with the different sizes of UIs, we normalize $x$ and $y$ relative to the width and height of the UI, respectively.

\subsubsection{Segment embedding} Segment embedding indicates whether the corresponding UI component is from UI-A or UI-B, and is a text or vision component. There are four types of segment embedding representing UI-A text, UI-B text, UI-A vision and UI-B vision, respectively. In practice, we define a fifth segment type, padding segment, to pad the input sequences to a fixed-length for batch processing.\\

The four types of input features are processed through a linear layer followed by a normalization layer \cite{ba2016layer}. Then they are summed up and passed as input to ActionBert, as a single tensor of shape $L*D_1$, where $L$ is the number of components in the UI pair and $D_1$ is the input embedding dimension. ActionBert is a uni-stream architecture, allowing attention across components and modalities. The output of ActionBert is a contextual embedding of shape $L*D_2$, where $D_2$ is the output embedding dimension of the ActionBert. The output embedding at position $i$ represents the contextual embedding of UI component $i$, while the embedding of the first component [CLS] provides an overall representation of the UI pair.

\subsection{Pre-training ActionBert} \label{sec:actionbert:pretrain}

ActionBert is pre-trained on three new tasks that are specifically designed to integrate user actions and UI-specific features: link component prediction, consecutive UI prediction, and masked VH text prediction. The first two pre-training tasks use UI sequences and user actions to learn the connectivity and relationship of two UIs. The last pre-training task learns the relationship between the text features of a UI component and its context (vision and text).

For pre-training, we used a large scale internal dataset obtained by automatically crawling various apps. Our data consists of 60,328 user action sequences on UIs. Each sequence $S$=[$s_1$, $s_2$...$s_T$] contains $T$ UIs, where $T$ ranges from two (a single click) to hundreds. Each pair of consecutive UIs ($s_{i-1}$, $s_i$) also has an action location ($x$,$y$), indicating the click position that results in the transition from $s_{i-1}$ to $s_i$. We extract 2.69M UIs with their view hierarchy from the sequences. We perform a 50\%-50\% negative sampling to generate non-consecutive UI pairs for the consecutive UI prediction task. Among the negative pairs, half are from the same sequence but not consecutive, while the other half of the negative pairs are from different user sequences. In total, the ActionBert is pre-trained on 5.4M UI pairs with user actions and view hierarchy.

\bheading{Pre-training task \#1: Link component prediction (LCP)} This task is specifically designed to incorporate the user action information from UI sequences. Given two UIs, the task is to predict which component can be clicked on to jump from UI-A to UI-B. The correct link component is obtained via user click position (x,y) during the training data generation. To correctly identify the link components, the model has to learn the semantics of both UIs and find a component whose functionality is to link them. The model takes all text and vision components of both UIs as candidates and selects one from them. The objective can be formulated as
{\small
\begin{linenomath}
\begin{align}
p&=softmax(MLP(f_{\theta}(x))),\\
L_{LCP}&=-\Sigma_{x \in D} \textbf{1}_{LC}(x)CE(p, y),
\end{align}
\end{linenomath}
}%
where $x$ is sampled from the training set $D$. $f_{\theta}(x)$ represents the embedding generated by the ActionBert model. $MLP(\cdot)$ is a multi-layer perceptron, and $p$ is the predicted probability of each UI component being the link component. $\textbf{1}_{LC}(x)$ is an indicator function whose value is 1 if the link component is available in this training pair, i.e. the two UIs are consecutive and the click location (x,y) refers to a valid UI component, otherwise 0. $CE(\cdot)$ is a standard multi-class cross-entropy loss and $y$ indicates the one-hot label of the correct link component.

\bheading{Pre-training task \#2: Consecutive UI prediction (CUI)} Inspired by the next sentence prediction task in BERT pre-training to model the relationship of two sentences, we propose this task to learn the relationship between two UIs. As shown in Table \ref{tab:concepts-map}, we analogize a UI to a sentence in NLP. The consecutive UIs prediction task predicts whether UI-B can be obtained by a single interaction from UI-A. In pre-training, a UI pair ($s_{i-1}$, $s_i$) from the same sequence $S$ is a positive training sample pair. We perform a 50\%-50\% negative sampling to generate negative samples. Among the negative samples, half of them (25\% of total training pairs) are generated by sampling two non-consecutive UIs from the same sequence, i.e. ($s_i, s_j$) where $i+1 \neq j$. The other half consists of two UIs from different user interaction sequences, i.e. ($s_i$, $v_j$), where $s_i$ is from sequence $S$ and $v_j$ is from sequence $V$ and $S \neq V$. Formally, the loss function is
{\small
\begin{linenomath}
\begin{align}
L_{CUI}=-\Sigma_{x \in D} ylog(\hat{y}) + (1-y)log(1-\hat{y})
\end{align}
\end{linenomath}
} %
where $x$ is a training sample (a pair of UIs) from the training set $D$, and $y$ is the label of whether the two UIs are consecutive. $\hat{y}$=sigmoid(MLP($f_{\theta}(x)$)) is the model predicted probability that the pair of UIs in $x$ are consecutive. A standard binary cross-entropy loss is applied to it.

\bheading{Pre-training task \#3: Masked VH text prediction.} This task is similar to the masked language modeling (MLM) task in BERT pre-training. We randomly mask $15\%$ text components from the UI view hierarchy. The main difference is, as each text token in ActionBert is a concatenation of multiple fields (content description, source id, component type and text) from the view hierarchy (Section \ref{sec:bg}), it contains more than one word token and we treat it as a text ``sentence''. Therefore, compared to BERT where each word token is directly predicted, ActionBert predicts the high-dimensional text embedding of the sentence and treats it as a regression task. Formally, the loss for masked VH text prediction is:
{\small
\begin{linenomath}
\begin{align}
L_{mask}=\Sigma_{x \in D} \Sigma_{i=0}^{N-1} \textbf{1}_{mask}(x,i)||f_{\theta}^{i}(x_{mask})-g^{i}(x)||_2^2
\end{align}
\end{linenomath}
} %
where $D$ is the training set. $x$ is the unmasked training example and $x_{mask}$ is the training example with masked text. $N$ is the total number of UI components in a training example. $f_\theta$ is the ActionBert model with parameter $\theta$, and $g$ is the sentence-level text encoder (we choose BERT in the pre-training), respectively. $f_{\theta}^{i}(x_{mask})$ denotes the ActionBert model output embedding of the $i$-th component in the masked example $x_{mask}$. $g^{i}(x)$ denotes the sentence encoder output of the $i$-th component in the unmasked example $x$. $\textbf{1}_{mask}(x,i)$ is an indicator function whose value is 1 if the component $i$ is masked in example $x$, otherwise 0. 

The overall loss function for pre-training is defined as:
{\small
\begin{linenomath}
\begin{align}
L=L_{LCP} + \lambda_{CUI} L_{CUI} + \lambda_{mask} L_{mask} \label{eq:loss-total}
\end{align}
\end{linenomath}
}%

\subsection{Fine-tuning ActionBert}

Similar to how BERT is used as a generic feature representation for different NLP downstream tasks, fine-tuning ActionBert for a variety of UI understanding tasks is relatively easy and does not require substantial task-specific architecture changes nor a large amount of task-specific data. The downstream input to ActionBert needs to be appropriately formatted into segments, as illustrated in Figure \ref{fig:actionbert:architecture}. During fine-tuning, a task-specific loss function is added on top of ActionBert for training. All parameters, including the text and vision encoder, are jointly tuned in an end-to-end manner to achieve the best performance.

Same as pre-training, we use VH texts and bounding boxes if VH is available in a downstream task. Otherwise, we perform OCR and object detection to extract the text and vision components from the UI, respectively. It is worth noting that, although the ActionBert is pre-trained on UI pairs, it can also handle single-UI and multi-UI tasks.

\subsubsection{Single-UI tasks}

As discussed above, the input data format is designed as [UI-A text, UI-B text, UI-A vision, UI-B vision]. For the downstream tasks which only involve a single UI, the input data can be converted into the ActionBert format by leaving the UI-B text and UI-B vision segments empty. For tasks involving natural language input, e.g., UI component retrieval based on its description, the language input can be passed in the UI-A text segment as a text token with the whole UI screenshot being used as the corresponding vision token.

\subsubsection{Two-UI tasks} 

Since the ActionBert model is pre-trained on UI pairs, it is natural to apply this model on tasks with two UIs as input, e.g. similar UI component retrieval (Section \ref{sec:exp:fine-tune}). We can assign the corresponding text and vision components to UI-A/B text and vision segments in Figure \ref{fig:actionbert:architecture}.

\subsubsection{Extension to multi-UI tasks} ActionBert can also be extended to multi-UI ($\ge$ 3) settings, though these types of tasks are not common in practice. Similar to two-UI tasks, different UIs and modalities need to be separated by the [SEP] token. The only difference is that more segment embedding representing the newer UI segments needs to be trained.
\section{Experiments} \label{sec:exp}

\subsection{Pre-training}

As described in Section \ref{sec:actionbert:pretrain}, we pre-train the ActionBert model on large-scale UI sequences with user actions and view hierarchy. We obtained the pre-training action sequences using the Robo app crawler \cite{robodocs}. In total, we used 60,328 UI sequences and extracted 5.4M UI pairs for pre-training. We split this data in the ratio of 80\%:10\%:10\% to obtain the train, dev and test sets, respectively. We prevent leakage of data across these splits by ensuring any app can go into only one of these splits. We pre-train two models of different sizes for comparison: ActionBert$_{Base}$ and ActionBert$_{Large}$ are 6-layers and 12-layers transformer architectures both with 6 heads and 768 hidden dimensions initialized with Glorot initialization \cite{glorot2010understanding}.

We use Adam optimizer \cite{kingma2014adam} with learning rate $r=10^{-5}$, $\beta_1=0.9$, $\beta_2=0.999$, $\epsilon=10^{-7}$ and batch size = 128 for training. We set $\lambda_{CUI}=0.1$ and $\lambda_{mask}=0.01$ in Eq. (\ref{eq:loss-total}) during the pre-training. ActionBert is pre-trained with 16 TPUs for three days.

\subsection{Fine-tuning} \label{sec:exp:fine-tune}

ActionBert can be fine-tuned on multiple types of downstream tasks. We evaluate it on four representative UI downstream tasks: similar component retrieval (across app and web UIs), referring expression component retrieval, icon classification and app type classification. Additionally, we perform link component prediction, one of the pre-training tasks, on a different dataset. To understand the performance of ActionBert, we use a benchmark model where we obtain embedding for each UI component by using ResNet to encode the image and BERT to encode the text attributes. The final classification layer added on top of the embedding is same as that added for ActionBert. Additionally, for each of the downstream tasks, we also evaluate the performance of a non pre-trained ActionBert model. This model has the same architecture as a pre-trained model but the model parameters are initialized randomly for each task. This comparison allows us to understand the impact of the transformer architecture and the pre-training tasks. Please refer to the appendix for more details regarding the different downstream tasks and the data collection details.

\subsubsection{Similar UI component retrieval}

\begin{figure}[h]
    \centering
    \includegraphics[width=0.7\columnwidth]{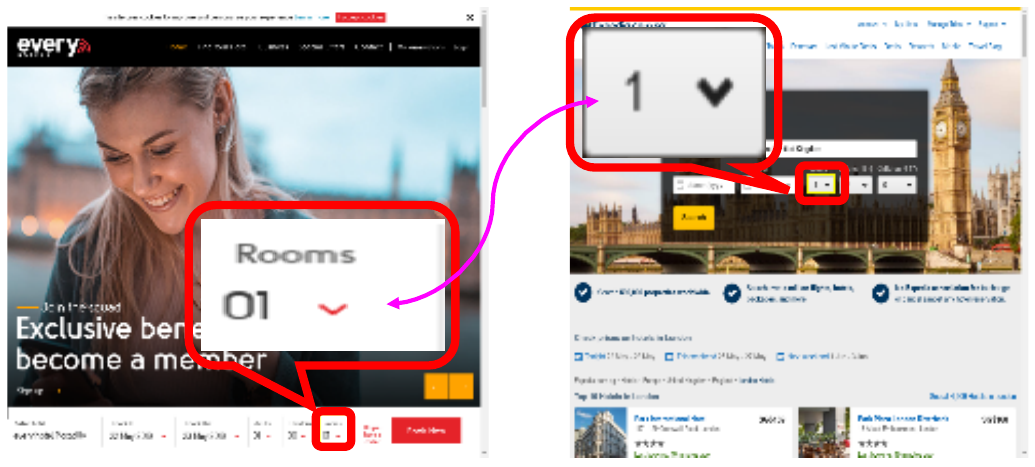}
    \caption{An example of similar UI component retrieval.}\label{fig:exp:sim-comp-web}
\end{figure}

Similar UI component retrieval focuses on the high-level functionality of UI components. Given an anchor component on a UI, the task is to choose the most similar component based on their functionality on the other UI from a list of UI component candidates (Figure \ref{fig:exp:sim-comp-web}). After generating the component-level embedding using ActionBert, we use dot-product of the embedding of the anchor and candidate components as the similarity scores and select the component with the highest score as the prediction. We use concatenated embeddings from ResNet + BERT for each UI component as the benchmark.

We perform similar UI component retrieval on app and web UIs. For the app UI task, we collected 900k app UI pairs with annotations of similar components for training, and 32k for validation and testing, respectively. \footnote{The view hierarchy is not available in this dataset, so we rely on OCR text from the screenshot instead.} For the web UI task, we collected 65k web UI pairs with annotations of similar components, which are jointly used with the app UIs for training. We manually labeled 2k web UI pairs with similar components for testing. On average, the model chooses the correct component from 10 candidate components on app UIs, and 35 candidates on web UIs.

\begin{table}[h]
\centering
\caption{Comparison to the baseline models for similar UI component retrieval on app and web UIs.} \label{tab:exp:sim-comp}
\resizebox{\linewidth}{!}{
\begin{tabular}{l|c|c|c|c}
\hline
\multicolumn{1}{c|}{}                       & \multicolumn{2}{c|}{App UI component Retrieval} & \multicolumn{2}{c}{Web UI component Retrieval} \\ \hline
\multicolumn{1}{c|}{Model}                  & Accuracy                 & Gain                 & Accuracy                & Gain                  \\ \hline
\multicolumn{1}{c|}{ResNet + BERT baseline} & 83.37                    & 0.00                 & --                      & --                    \\ \hline
MobileNet + USE baseline                    & --                       & --                   & 50.16                   & 0.00                  \\ \hline
ActionBert$_{BASE-NP}$                       & 85.18                    & +1.81                & 62.85                   & +12.69                \\ \hline
ActionBert$_{BASE}$                          & 85.53                    & +2.16                & 63.67                   & +13.51                \\ \hline
ActionBert$_{LARGE-NP}$                      & 86.13                    & +2.76                & 62.14                   & +11.98                \\ \hline
ActionBert$_{LARGE}$                         & \textbf{86.43}           & \textbf{+3.06}       & \textbf{64.41}          & \textbf{+14.25}       \\ \hline
\end{tabular}
}
\end{table}

Table \ref{tab:exp:sim-comp} shows the results of similar UI component retrieval. For the app similar component retrieval, the pre-trained ActionBert$_{LARGE}$ model outperforms the baseline ResNet+BERT model by 3.06\% in accuracy. For the web similar component retrieval task, the pre-trained model outperforms the benchmark MobileNet + Universal Sentence Encoder (USE) \cite{cer2018universal} model, which is specifically designed for web similar component retrieval, by 14.25\%. Furthermore, the pre-trained models (ActionBert$_{BASE/LARGE}$) achieve higher accuracy than the non pre-trained models (ActionBert$_{BASE/LARGE-NP}$), which shows the benefit of using a generic pre-trained feature representation in UI understanding.

\subsubsection{Referring expression component retrieval}

\begin{figure}[h]
    \centering
    \includegraphics[width=0.7\columnwidth]{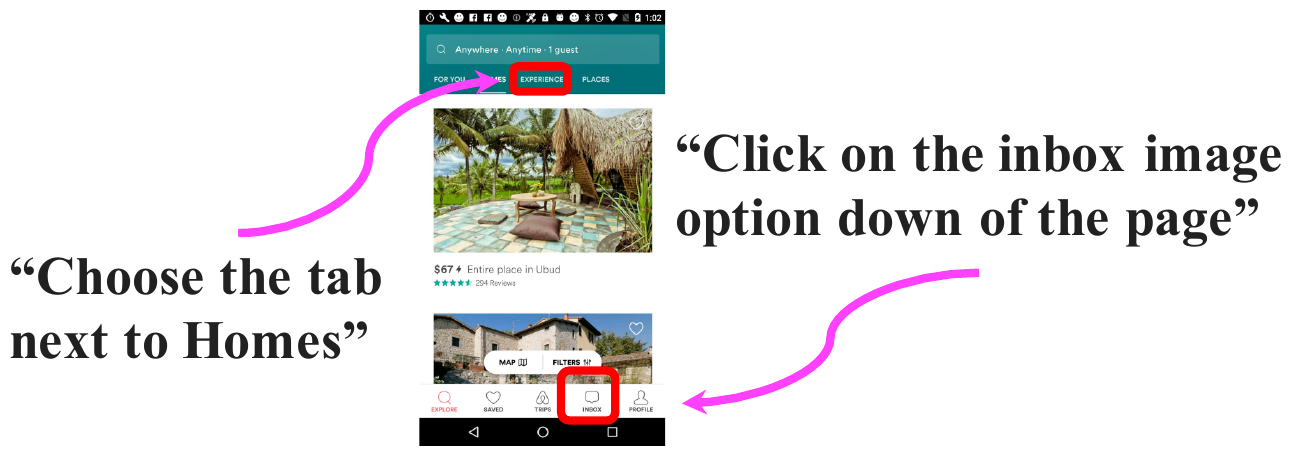}
    \caption{Referring expression UI component retrieval task.}\label{fig:exp:ref-exp}
\end{figure}

The referring expression component retrieval task takes a referring expression and a UI screenshot as input, and the model is expected to select the UI component that the expression refers to (Figure \ref{fig:exp:ref-exp}) from a list of the UI components detected on the screen. This is a typical task in voice-based control systems where the user can interact with an element on the screen by describing it, e.g., ``click settings'', ``select on the back arrow at the top''. To correctly select the UI component, the model not only has to take into account the semantics of the component, but also its relationship to other components.

We collected and manually annotated 16.9k UI components with referring expressions for training, 2.1k for validation and 1.8k for testing. We used a fine-tuned Faster-RCNN to select negative samples (components that are not referred by the expression) in a UI, by selecting the objects with high object detection scores that do not overlap with the correct component. On average, the model needs to choose the correct component that the expression refers to from 20 UI components.

\begin{table}[h]
\centering
\caption{Comparison to the baseline ResNet+BERT embedding for UI referring expression component retrieval.} \label{tab:exp:ref-exp}
\resizebox{0.7\linewidth}{!}{
\begin{tabular}{l|c|c|c}
\hline
\multicolumn{1}{c|}{Model} & Dev   & Test           & Test Gain      \\ \hline
ResNet+BERT baseline        & 87.62 & 86.19          & --             \\ \hline
ActionBert$_{BASE-NP}$      & 86.71 & 85.68          & -0.51          \\ \hline
ActionBert$_{BASE}$         & 90.14 & 88.38          & +2.19          \\ \hline
ActionBert$_{LARGE-NP}$     & 89.54 & 89.17          & +2.96          \\ \hline
ActionBert$_{LARGE}$        & 89.84 & \textbf{90.16} & \textbf{+3.97} \\ \hline
\end{tabular}
}
\end{table}

We present the results of referring expression UI component retrieval in Table \ref{tab:exp:ref-exp}. We observe that the pre-trained ActionBert$_{LARGE}$ model performs the best, and achieves 3.97\% improvement over the ResNet+BERT benchmark model. Moreover, the pre-trained models  (ActionBert$_{BASE/LARGE}$) outperforms the non pre-trained models  (ActionBert$_{BASE-NP/LARGE-NP}$).

\subsubsection{Icon classification}

The goal of this task is to identify the type of an icon. Having this information is beneficial for screen readers to understand the type of elements when content description and alt-text are not present. We use the Rico \cite{deka2017rico} dataset for this task. Rico is the largest public mobile app design dataset, containing 72k unique screenshots with their view hierarchies. However, the icons in the dataset are labeled using heuristics and simple ML models relying on view hierarchy attributes making the annotations inaccurate. Hence, we use crowd-sourcing to label icons of this dataset in two degrees of granularity, i.e. 32 top-used icon classes and 77 more detailed icon classes.

\begin{table}[h]
\caption{Comparison to the baseline ResNet+BERT embedding for icon classification.} \label{tab:exp:icon-cls}
\resizebox{\linewidth}{!}{
\begin{tabular}{l|l|l|l|l}
\hline
                        & \multicolumn{2}{c|}{Rico-32 Classes}    & \multicolumn{2}{c}{Rico-77 Classes}    \\ \hline
Model                   & Micro Accuracy   & Macro F1        & Micro Accuracy   & Macro F1        \\ \hline
ResNet+BERT baseline    & 97.32          & 0.8655          & 91.28          & 0.6256          \\ \hline
ActionBert$_{BASE-NP}$  & 97.38          & 0.8667          & 91.57          & 0.6303          \\ \hline
ActionBert$_{BASE}$     & 97.42          & 0.8742          & 91.60          & \textbf{0.6376}          \\ \hline
ActionBert$_{LARGE-NP}$ & 97.39          & 0.8703          & 91.56          & 0.6317          \\ \hline
ActionBert$_{LARGE}$    & \textbf{97.50} & \textbf{0.8766} & \textbf{91.65} & 0.6307 \\ \hline
\end{tabular}
}
\end{table}

We use the ActionBert embedding in the corresponding position as the contextual embedding of the UI components for icon classification. We summarize the icon classification results in Table \ref{tab:exp:icon-cls}. For icon-32 classification, the pre-trained ActionBert$_{LARGE}$ model achieves the best macro-accuracy of 97.50\% and micro-F1 of 0.8766, 0.18\% and 1.11\% higher than the ResNet+BERT benchmark, respectively. On the finer-granularity (77 classes) icon classification, ActionBert$_{LARGE}$ obtains 0.37\% improvement on accuracy over the baseline. A smaller model ActionBert$_{BASE}$ performs slightly better on the micro-F1 metric, because of the skewed class distribution in the RICO dataset.

\subsubsection{App type classification}

In this task, the goal of the model is to predict the type of an app, e.g., shopping, communication, arts etc. In total, we examined 27 app types. Similar to icon classification, we use the public Rico dataset for this task. The app type is extracted from the description in the app store. We use 43.5k unique app UIs with their view hierarchies and app types, and split them in the ratio 80\%, 10\%, 10\% for training, validation and testing. Compared to the icon classification above, which is a component-level task, the app type classification is a UI-level task.

\begin{table}[h]
\centering
\caption{Comparison to the baseline ResNet+BERT embedding for app type classification.} \label{tab:exp:app-type}
\resizebox{0.9\linewidth}{!}{
\begin{tabular}{l|c|c|c|c}
\hline
\multicolumn{1}{c|}{Model} & Micro accuracy & Gain          & Macro F1       &  Gain    \\ \hline
ResNet+BERT baseline        & 64.3          & --            & 0.598          &  --      \\ \hline
ActionBert$_{BASE-NP}$      & 78.6          & 14.3          & 0.753          &  15.5    \\ \hline
ActionBert$_{BASE}$         & \textbf{79.8} & \textbf{15.5} & \textbf{0.764} & \textbf{16.6} \\ \hline
\end{tabular}
}
\end{table}

From Table \ref{tab:exp:app-type} we can see that Compared to the ResNet+BERT benchmark model, the pre-trained ActionBert leads to an increase in accuracy by 15.5\% and macro F1 by 16.6\%. Furthermore, pre-trained ActionBert outperforms the non pre-trained version by 1.2\% and 1.1\% on micro accuracy and macro F1 metrics, respectively. This demonstrates the effectiveness of the pre-training tasks.

\subsubsection{Link component prediction}

We also evaluate the performance of ActionBert on link component prediction (Section \ref{sec:actionbert:pretrain}), one of the pre-training tasks, on the publicly available Rico dataset. Rico also contains 10k user interaction sequences, but typically the length of these sequences is less than the ones we used for pre-training. We extract 90k consecutive UI pairs from the Rico sequences and perform link component prediction on them.

\begin{table}[h]
\centering
\caption{Comparison to the baseline ResNet+BERT embedding on link component prediction (a pre-training task).} \label{tab:exp:lc}
\resizebox{0.6\linewidth}{!}{
\begin{tabular}{l|c|c}
\hline
                       & \multicolumn{1}{l|}{Accuracy} & \multicolumn{1}{l}{Gain} \\ \hline
ResNet+BERT baseline   & 40.2                          & --                        \\ \hline
ActionBert$_{BASE-NP}$ & 48.2                          & +8.0                       \\ \hline
ActionBert$_{BASE}$    & \textbf{51.6}                 & \textbf{+11.4}             \\ \hline
\end{tabular}
}
\vspace{-10pt}
\end{table}

From Table \ref{tab:exp:lc}, we can see that Pre-trained and non pre-trained ActionBert models significantly outperform the baseline ResNet+BERT model, showing the benefit of the unified single-stream attention architecture. The pre-trained ActionBert model outperforms the non pre-trained one by 3.4 \%. 

In summary, across all the fine-tuning tasks, the pre-trained ActionBert outperforms the baselines and the non pre-trained model, suggesting the effectiveness of our proposed method in generating generic UI embedding.
\section{Related Work} \label{sec:related}

Previous works have studied UI embeddings for specific applications, e.g., UI search and design. \cite{deka2017rico} and \cite{huang2019swire} proposed using auto-encoder and VGG-style architectures to embed UI layout information for similar UI retrieval, respectively. \cite{liu2018learning} used also finer element-level layout information for retrieval. \cite{wichers2018resolving} demonstrated an embedding approach to retrieve images from a referring expression. Besides UI search, UI understanding provides insights to the UI designers, e.g., predicting user engagement level \cite{wu2020predicting}, user impressions of the app \cite{wu2019understanding} and perceived functionality of the UI element \cite{swearngin2019modeling}. These works each targeted a specific UI task, while we are the first to show that a generic pre-trained feature representation can help improve various UI understanding tasks. Furthermore, none of the previous works have investigated user actions to improve UI understanding.



Concurrently, beyond the scope of UI understanding, research work has been proposed on pre-training feature representation for vision-linguistic tasks, e.g., VL-BERT \cite{su2019vl}, ViL-BERT \cite{lu2019vilbert}, B2T2 \cite{alberti2019fusion}, VisualBERT \cite{li2019visualbert}, UNITER \cite{chen2020uniter} and ImageBERT \cite{qi2020imagebert}. However, these work focus on natural language and images. They do not consider the domain-specific user actions and UI features, thus are not directly applicable to UI understanding. Our proposed model explicitly integrates UI domain-specific information into the pre-training process.

\section{Conclusion} \label{sec:conclusion}

In this paper, we explore using user actions to build generic feature representations to facilitate UI understanding. We present ActionBert, the first pre-trained UI embedding model that can be applied to multiple UI understanding tasks. ActionBert is pre-trained on a large dataset of sequential user actions and UI domain-specific features. Experiment results show that pre-training helps in improving the performance across four types of representative UI tasks. ActionBert also significantly outperforms multi-modal baselines across all downstream tasks by up to 15.5\%. We hope that this study can raise awareness about the importance of pre-trained feature representations in this field and spur the development of useful models for various UI related tasks.

\section*{Acknowledgements}
We thank Hakim Sidahmed, Harrison Lee, Raghav Gupta and anonymous reviewers for reviewing the manuscript and providing valuable feedback; Abhinav Rastogi and James Stout for ideas about UI embedding which inspired our work; Maria Wang for her guidance and help on dataset creation; Chongyang Bai, Xinying Song and Hao Zhang for their insightful discussion and feedback; Pranav Khaitan for guidance and encouragement.

\bibliography{main}

\begin{thebibliography}{22}
\providecommand{\natexlab}[1]{#1}
\providecommand{\url}[1]{\texttt{#1}}
\providecommand{\urlprefix}{URL }
\expandafter\ifx\csname urlstyle\endcsname\relax
  \providecommand{\doi}[1]{doi:\discretionary{}{}{}#1}\else
  \providecommand{\doi}{doi:\discretionary{}{}{}\begingroup
  \urlstyle{rm}\Url}\fi

\bibitem[{Alberti et~al.(2019)Alberti, Ling, Collins, and
  Reitter}]{alberti2019fusion}
Alberti, C.; Ling, J.; Collins, M.; and Reitter, D. 2019.
\newblock Fusion of Detected Objects in Text for Visual Question Answering.
\newblock \emph{arXiv preprint arXiv:1908.05054} .

\bibitem[{Ba, Kiros, and Hinton(2016)}]{ba2016layer}
Ba, J.~L.; Kiros, J.~R.; and Hinton, G.~E. 2016.
\newblock Layer Normalization.
\newblock \emph{arXiv preprint arXiv:1607.06450} .

\bibitem[{Cer et~al.(2018)Cer, Yang, Kong, Hua, Limtiaco, John, Constant,
  Guajardo-Cespedes, Yuan, Tar et~al.}]{cer2018universal}
Cer, D.; Yang, Y.; Kong, S.-y.; Hua, N.; Limtiaco, N.; John, R.~S.; Constant,
  N.; Guajardo-Cespedes, M.; Yuan, S.; Tar, C.; et~al. 2018.
\newblock Universal Sentence Encoder.
\newblock \emph{arXiv preprint arXiv:1803.11175} .

\bibitem[{Chen et~al.(2020)Chen, Li, Yu, Kholy, Ahmed, Gan, Cheng, and
  Liu}]{chen2020uniter}
Chen, Y.-C.; Li, L.; Yu, L.; Kholy, A.~E.; Ahmed, F.; Gan, Z.; Cheng, Y.; and
  Liu, J. 2020.
\newblock Uniter: Universal Image-text Representation Learning.
\newblock In \emph{European Conference on Computer Vision (ECCV)}.

\bibitem[{Deka et~al.(2017)Deka, Huang, Franzen, Hibschman, Afergan, Li,
  Nichols, and Kumar}]{deka2017rico}
Deka, B.; Huang, Z.; Franzen, C.; Hibschman, J.; Afergan, D.; Li, Y.; Nichols,
  J.; and Kumar, R. 2017.
\newblock Rico: A Mobile App Dataset for Building Data-driven Design
  Applications.
\newblock In \emph{ACM Symposium on User Interface Software and Technology
  (UIST)}.

\bibitem[{Devlin et~al.(2018)Devlin, Chang, Lee, and
  Toutanova}]{devlin2018bert}
Devlin, J.; Chang, M.-W.; Lee, K.; and Toutanova, K. 2018.
\newblock Bert: Pre-training of Deep Bidirectional Transformers for Language
  Understanding.
\newblock \emph{arXiv preprint arXiv:1810.04805} .

\bibitem[{Firebase(2020)}]{robodocs}
Firebase. 2020.
\newblock Robo App Crawler Documentation.
\newblock \url{https://firebase.google.com/docs/test-lab/android/robo-ux-test}.

\bibitem[{Glorot and Bengio(2010)}]{glorot2010understanding}
Glorot, X.; and Bengio, Y. 2010.
\newblock Understanding the Difficulty of Training Deep Feedforward Neural
  Networks.
\newblock In \emph{International Conference on Artificial Intelligence and
  Statistics (AISTATS)}.

\bibitem[{He et~al.(2016)He, Zhang, Ren, and Sun}]{he2016deep}
He, K.; Zhang, X.; Ren, S.; and Sun, J. 2016.
\newblock Deep Residual Learning for Image Recognition.
\newblock In \emph{IEEE Conference on Computer Vision and Pattern Recognition
  (CVPR)}.

\bibitem[{Huang, Canny, and Nichols(2019)}]{huang2019swire}
Huang, F.; Canny, J.~F.; and Nichols, J. 2019.
\newblock Swire: Sketch-Based User Interface Retrieval.
\newblock In \emph{ACM CHI Conference on Human Factors in Computing Systems
  (CHI)}.

\bibitem[{Kingma and Ba(2014)}]{kingma2014adam}
Kingma, D.~P.; and Ba, J. 2014.
\newblock Adam: A Method for Stochastic Optimization.
\newblock \emph{arXiv preprint arXiv:1412.6980} .

\bibitem[{Li et~al.(2019)Li, Yatskar, Yin, Hsieh, and Chang}]{li2019visualbert}
Li, L.~H.; Yatskar, M.; Yin, D.; Hsieh, C.-J.; and Chang, K.-W. 2019.
\newblock Visualbert: A Simple and Performant Baseline for Vision and Language.
\newblock \emph{arXiv preprint arXiv:1908.03557} .

\bibitem[{Liu et~al.(2018)Liu, Craft, Situ, Yumer, Mech, and
  Kumar}]{liu2018learning}
Liu, T.~F.; Craft, M.; Situ, J.; Yumer, E.; Mech, R.; and Kumar, R. 2018.
\newblock Learning Design Semantics for Mobile Apps.
\newblock In \emph{ACM Symposium on User Interface Software and Technology
  (UIST)}.

\bibitem[{Lu et~al.(2019)Lu, Batra, Parikh, and Lee}]{lu2019vilbert}
Lu, J.; Batra, D.; Parikh, D.; and Lee, S. 2019.
\newblock Vilbert: Pretraining Task-agnostic Visiolinguistic Representations
  for Vision-and-language Tasks.
\newblock In \emph{Advances in Neural Information Processing Systems
  (NeurIPS)}.

\bibitem[{Qi et~al.(2020)Qi, Su, Song, Cui, Bharti, and
  Sacheti}]{qi2020imagebert}
Qi, D.; Su, L.; Song, J.; Cui, E.; Bharti, T.; and Sacheti, A. 2020.
\newblock Imagebert: Cross-modal Pre-training with Large-scale Weak-supervised
  Image-text Data.
\newblock \emph{arXiv preprint arXiv:2001.07966} .

\bibitem[{Ren et~al.(2015)Ren, He, Girshick, and Sun}]{ren2015frcnn}
Ren, S.; He, K.; Girshick, R.; and Sun, J. 2015.
\newblock Faster R-CNN: Towards Real-time Object Detection with Region Proposal
  Networks.
\newblock In \emph{Advances in neural information processing systems
  (NeurIPS)}.

\bibitem[{Su et~al.(2019)Su, Zhu, Cao, Li, Lu, Wei, and Dai}]{su2019vl}
Su, W.; Zhu, X.; Cao, Y.; Li, B.; Lu, L.; Wei, F.; and Dai, J. 2019.
\newblock VL-BERT: Pre-training of Generic Visual-Linguistic Representations.
\newblock In \emph{International Conference on Learning Representations
  (ICLR)}.

\bibitem[{Swearngin and Li(2019)}]{swearngin2019modeling}
Swearngin, A.; and Li, Y. 2019.
\newblock Modeling Mobile Interface Tappability Using Crowdsourcing and Deep
  Learning.
\newblock In \emph{ACM CHI Conference on Human Factors in Computing Systems
  (CHI)}.

\bibitem[{Vaswani et~al.(2017)Vaswani, Shazeer, Parmar, Uszkoreit, Jones,
  Gomez, Kaiser, and Polosukhin}]{vaswani2017attention}
Vaswani, A.; Shazeer, N.; Parmar, N.; Uszkoreit, J.; Jones, L.; Gomez, A.~N.;
  Kaiser, {\L}.; and Polosukhin, I. 2017.
\newblock Attention Is All You Need.
\newblock In \emph{Advances in Neural Information Processing Systems
  (NeurIPS)}.

\bibitem[{Wichers, Hakkani-T{\"u}r, and Chen(2018)}]{wichers2018resolving}
Wichers, N.; Hakkani-T{\"u}r, D.; and Chen, J. 2018.
\newblock Resolving Referring Expressions in Images with Labeled Elements.
\newblock In \emph{IEEE Spoken Language Technology Workshop (SLT)}.

\bibitem[{Wu et~al.(2020)Wu, Jiang, Liu, and Ma}]{wu2020predicting}
Wu, Z.; Jiang, Y.; Liu, Y.; and Ma, X. 2020.
\newblock Predicting and Diagnosing User Engagement with Mobile UI Animation
  via a Data-Driven Approach.
\newblock In \emph{ACM CHI Conference on Human Factors in Computing Systems
  (CHI)}.

\bibitem[{Wu et~al.(2019)Wu, Kim, Li, and Ma}]{wu2019understanding}
Wu, Z.; Kim, T.; Li, Q.; and Ma, X. 2019.
\newblock Understanding and Modeling User-Perceived Brand Personality from
  Mobile Application UIs.
\newblock In \emph{ACM CHI Conference on Human Factors in Computing Systems
  (CHI)}.

\end{thebibliography}

\end{document}